\documentclass{article}
\usepackage{ijcai22}

\usepackage{times}
\usepackage{soul}
\usepackage{url}
\usepackage[hidelinks]{hyperref}
\usepackage[utf8]{inputenc}
\usepackage[small]{caption}
\usepackage{graphicx}
\usepackage{amsmath}
\usepackage{amsthm}
\usepackage{booktabs}
\usepackage{algorithm}
\usepackage{algorithmic}
\usepackage{amssymb}
\usepackage{mathtools}      
\usepackage{graphicx}       
\usepackage{appendix}       
\DeclareMathOperator*{\argmax}{arg\,max}
\DeclareMathOperator*{\argmin}{arg\,min}

\DeclareMathOperator{\xinput}{\mathbf{x}}

\title{How Does Frequency Bias Affect the Robustness of Neural Image Classifiers \\against Common Corruption and Adversarial Perturbations?}

\author{Alvin Chan$^{1,2}$\thanks{Corresponding author: \texttt{guoweialvin.chan@ntu.edu.sg}},~~ Yew-Soon Ong$^{2,1}$,~~ Clement Tan$^{1,2}$\\
$^1$Nanyang Technological University \:\: $^2$Agency for Science, Technology and Research, Singapore
}

\begin{document}

\maketitle

\begin{abstract}
Model robustness is vital for the reliable deployment of machine learning models in real-world applications. Recent studies have shown that data augmentation can result in model over-relying on features in the low-frequency domain, sacrificing performance against low-frequency corruptions, highlighting a connection between frequency and robustness. Here, we take one step further to more directly study the frequency bias of a model through the lens of its Jacobians and its implication to model robustness. To achieve this, we propose Jacobian frequency regularization for models' Jacobians to have a larger ratio of low-frequency components. Through experiments on four image datasets, we show that biasing classifiers towards low (high)-frequency components can bring performance gain against high (low)-frequency corruption and adversarial perturbation, albeit with a tradeoff in performance for low (high)-frequency corruption. Our approach elucidates a more direct connection between the frequency bias and robustness of deep learning models.\footnote{Full version with the Appendix in arXiv. Code available at:\\ \texttt{https://github.com/alvinchangw/JaFR\_IJCAI2022}}


\end{abstract}

\section{Introduction}
Recent research has shown that model performances can drop drastically when natural test images are altered \cite{szegedy2013intriguing,hendrycks2021natural,hendrycks2021many}. 
One of these scenarios is when common image corruptions are added to the images. These corruptions include noise attributed to weather conditions, noisy environments, blurring effects and digital artifacts \cite{hendrycks2019benchmarking}. Another case where models can fail is adversarial examples where a malicious party can craft imperceptible perturbations to images to influence the model's prediction \cite{szegedy2013intriguing,carlini2017towards,papernot2018cleverhans,croce2019minimally}. While research to build models more robust to these two situations started independently, recent studies have started to draw a connection between them. Interestingly, models trained to be robust against adversarial examples have shown mixed results in common corruptions: improving accuracy for some corruptions types while doing poorly for others.



\begin{figure}[!htbp]
    \centering
    \includegraphics[width=0.65\linewidth]{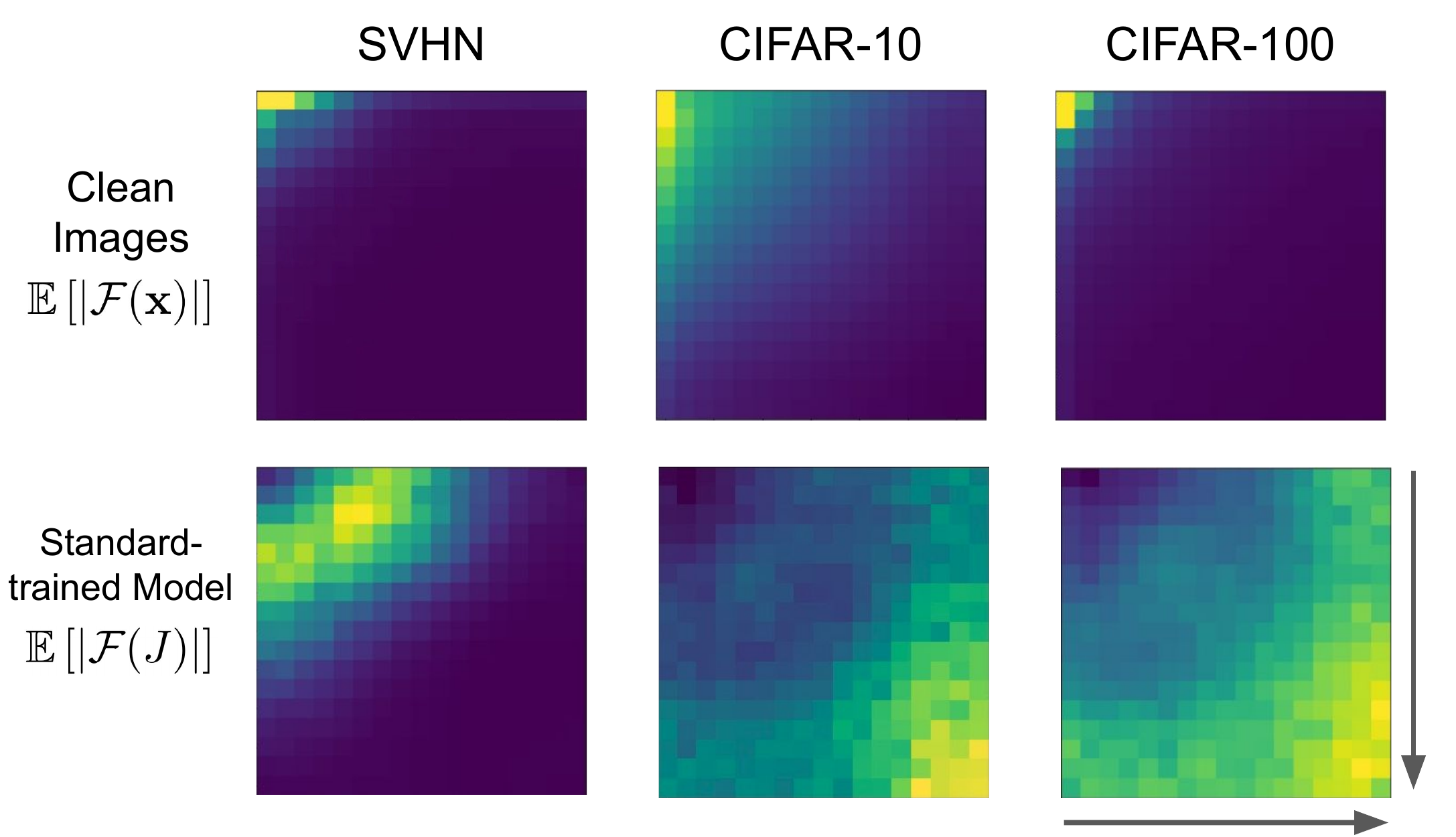}
    \caption{Fourier spectra of image datasets and the Jacobians of models trained on them, showing a mismatch between the frequency profiles of the images and models that are trained on them. Arrows show the direction of increasing frequency. Clean images contain large ratios of low-frequency components, shown by the intensity on the top left of the spectra (top row) while their corresponding standard-trained models rely heavily on high-frequency features (bottom row).}
    \label{fig:Fourier Spectra imgs and models}
\end{figure}

Though studies have shown a link between data augmentation strategies and robustness against corruptions with different frequency components \cite{yin2019fourier}, there is no study on how \emph{direct} changes to a model's Fourier profile would affect its robustness. Here, we aim to directly alter the Fourier profile of a model to study its direct effect on both the model's adversarial and corruption robustness. To achieve this, we investigate the model's Jacobian which represents a visual map of pixel importance in a particular input image \cite{smilkov2017smoothgrad}. Intuitively, a model would be relying on low (high)-frequency features when its Jacobians have a large ratio of low (high)-frequency components. While observing the Fourier spectra of natural images such as SVHN, CIFAR-10 and CIFAR-100, and the Jacobians of standard-trained models (Figure~\ref{fig:Fourier Spectra imgs and models}), these images have a much larger component of low-frequency features than their standard-trained models. This mismatch of frequency profiles motivates us to train models to bias towards low-frequency features through its Jacobians and study its effect on robustness. 

To quantify the frequency profile of a model, we propose a frequency bias term that computes a scalar value from a Fourier spectrum of 2-D inputs such as an image or Jacobian to improve frequency evaluation beyond the visual inspection of Fourier spectra. Through this differentiable frequency bias term, we can use Jacobian frequency regularization (JaFR) to explicitly train a model to bias more heavily on low- or high-frequency features. Through our experiments, we find that a more direct change in a model's frequency profile towards low-frequency regions to match the frequency profile of the training data can boost clean accuracy, adversarial robustness and common corruptions in certain settings while trading off performance against low-frequency noise. Conversely, regularizing the model towards high-frequency regions can boost performance against low-frequency noise while sacrificing accuracy under high-frequency noise and adversarial examples. Our main goal here is not to claim state-of-the-art robustness but to elucidate a more direct link between models' frequency characteristics and robustness against noise.
All in all, the core contributions of this paper are:
\begin{itemize}
    \item We propose a frequency bias term to measure the frequency bias of a Fourier spectrum.
    \item We show that biasing the Jacobians of models towards low or high frequency have implications on model robustness against adversarial robustness and an array of corruptions.
    \item To achieve this, we propose Jacobian frequency regularization (JaFR) to train model's Jacobians to have a larger or smaller weightage of low-frequency components.
    \item We conduct experiments on SVHN, CIFAR-10, CIFAR-100 and TinyImageNet to show how low-frequency bias in Jacobians can improve robustness against adversarial and high-frequency corruptions, albeit with tradeoffs in performance for low-frequency corruptions.
\end{itemize}

\section{Background and Related Work}


\textbf{Adversarial Robustness} \label{sec:adversarial robustness}
Adversarial robustness measures how well a model is resistant to attacks by malicious actors. In such attacks, imperceptible perturbations could be crafted to form adversarial examples with the aim to control the prediction of neural networks \cite{szegedy2013intriguing}. This threat could undermine the deployment of deep learning models in mission-critical applications.
In a classification task, a model ($f$) parameterized by $\theta$ takes an input $\xinput$ to predict the probabilities for $k$ classes, i.e., $f(\xinput ; \theta): \xinput \mapsto \mathbb{R}^k$. In supervised setting of empirical risk minimization (ERM), given training samples $(\xinput, \mathbf{y}) \sim D$, the model's parameters are trained to minmize the standard cross-entropy loss:
\begin{equation}
 \mathcal{L}(\xinput, \mathbf{y}) = \mathbb{E}_{(\xinput, \mathbf{y}) \sim D} \left[ - \mathbf{y}^\top \log f (\xinput) \right]
\end{equation}
where $\mathbf{y} \in \mathbb{R}^k$ is the one-hot label for the input $\xinput$.
Though ERM can train models that display high accuracy on a holdout test set, their performance degrades under adversarial examples. Given an adversarial perturbation of magnitude $\varepsilon$, we say a model is robust against this attack if

\begin{multline}
    \argmax_{i \in C} f_i(\xinput ; \theta) = \argmax_{i \in C} f_i(\xinput + \delta ; \theta) ~, \\~~~ \forall \delta \in B_p (\varepsilon) = {\delta : \| \delta \|_p \leq \varepsilon}
\end{multline}

where $p = \infty$ in this paper and is the most widely studied scenario.

One of the most effective approaches to train models robust against adversarial examples is adversarial training (AT) \cite{goodfellow2014explaining,madry2017towards,zhang2019defense,qin2019adversarial,andriushchenko2020understanding,wu2020adversarial}. More details of related work in adversarial training are deferred to the Appendix.

\textbf{Corruption Robustness} \label{sec:corruption robustness}
In contrast to adversarial robustness, corruption robustness \cite{hendrycks2019benchmarking} entails studying the performance of models when input data are corrupted with common-occurring noise, not necessarily those created by a malicious actor to control models' prediction. CIFAR-10-C and CIFAR-100-C are two benchmark datasets that study 19 corruption types, each with 5 levels of severity. These 19 corruption types can be grouped under the `Noise', `Blur', `Weather', `Digital' categories. There is a line of work that seek to improve performances under these common corruptions by mostly improving the training data augmentation \cite{geirhos2018imagenet,cubuk2018autoaugment,hendrycks2019augmix,lopes2019improving,hendrycks2020many,kapoor2020fourier,rusak2020simple,vasconcelos2020effective}. Some other works assemble expert models whose performance are finetuned for subsets of the corruptions \cite{lee2020compounding,saikia2021improving} to boost overall performance. Different from these work, our paper aims to study the effect of frequency bias on corruption robustness of different types that corrupts features of varying frequencies, rather than to propose a new way to better resist these corruptions.


\textbf{Link between Frequency and Robustness} \label{sec:fourier perspectives on robustness}
There is a line of work that seeks to understand how robust models respond to corruption and adversarial perturbations of various frequency profiles \cite{yin2019fourier,sharma2019effectiveness,ortiz2020hold,tsuzuku2019structural,vasconcelos2021impact}. Details of these works are in the Appendix. 
In contrast to these prior works, our work here takes the frequency analysis in a different direction by studying models through the Fourier spectrum of their Jacobians rather than their test or training data. More concretely, with the original training data, we train models and bias the frequency profile of the model's Jacobians towards low-frequency regions to see its effect on model robustness.

\textbf{Jacobians of Robust Models} \label{sec:jacobians of robust models}
The Jacobian, 
\begin{equation}
    J \coloneqq \nabla_{\xinput} \mathcal{L(\xinput, \mathbf{y})} 
\end{equation}
defines how the model's prediction changes with an infinitesimally small change to the input $\xinput$. For image classification, Jacobians can be loosely interpreted as a map of which pixels affect the model's prediction the most and, hence, give an illustration of important regions in an input image \cite{smilkov2017smoothgrad,adebayo2018sanity,etmann2019connection,ilyas2019adversarial}. There is a line of work that seeks to improve the adversarial robustness of models by matching the Jacobians to a target distribution \cite{chan2019jacobian,chan2020thinks} or by constraining their magnitude \cite{ross2018improving,jakubovitz2018improving}. Rather than aiming to improve the adversarial robustness, the core aim of our paper here is to investigate the relationship between the Fourier profile of models and robustness against corruptions. Moreover, the regularizing effect of JaFR has a different mechanism that acts directly on the Fourier spectrum of the Jacobians. A detailed comparison is in the Appendix.

\section{Jacobian Frequency Bias}
\textbf{Motivation} 
As mentioned in \S~\ref{sec:fourier perspectives on robustness}, it has been shown that there is a link between Fourier profile of input training data and robustness against corruptions with different frequency components. However, there is still no study on how changes to a neural network's Fourier profile would affect its robustness. Since the Jacobian of a model represents a visual map of pixel importance  \cite{smilkov2017smoothgrad}, it offers a medium for us to regularize the Fourier profile of the model. Intuitively, when a model's Jacobians are concentrated with low-frequency components, it places more importance on low-frequency features. Conversely, the model relies more on high-frequency features if its Jacobians have a relatively large proportion of high-frequency components. Here, we aim to study how changing the Fourier spectrum of the Jacobians would affect its robustness.

When analyzing the Fourier profile of Jacobians of adversarially robust models (Table~\ref{tab:cifar10 model fourier profile}), we see that it resembles the profile of the training data much more than the non-robust standard trained models. This raises the question of what would happen to model robustness if we directly train neural networks to have a low-frequency profile, similar to what we see in the images from SVHN, CIFAR-10 and CIFAR-100 (see Figure~\ref{fig:Fourier Spectra imgs and models}). This motivates a metric to quantitatively measure and control the Fourier profile of the neural network to more directly study the effect of Fourier profile on robustness. In the next sections, we propose the Jacobian frequency bias to achieve this goal and discuss how we can use it to train a neural network. 

\subsection{Jacobian Frequency Bias}
Here, we present the measure of Jacobian frequency bias with an example of single-channel image classification where the input images are denoted as $\xinput \in \mathbb{R}^{hw}$. Given a training dataset ($\mathcal{D}_{\text{train}}$) where each training sample consists of an input image $\xinput$ and one-hot label vector of $k$ classes as $\mathbf{y} \in \mathbb{R}^k$, we can express $f_{\text{cls}} (\xinput) \in \mathbb{R}^k$ as the prediction of the classifier ($f_{\text{cls}}$), parameterized by $\theta$. Then, the classification cross entropy loss ($\mathcal{L}_{\text{cls}}$) is:
\begin{equation} \label{eq:cls xent}
    \mathcal{L}_{\text{cls}} = - \mathbf{y}^\top \log f_{\text{cls}} (\xinput)
\end{equation}

where $f_{\text{cls}}$ is the classifier model. The Jacobian matrix $J \in \mathbb{R}^{hw}$ of the model's classication loss value with respect to the input layer can be computed through backpropagation:
\begin{equation}
    J(\xinput) \coloneqq \nabla_{\xinput} \mathcal{L}_{\text{cls}} = \begin{bmatrix}
  \frac{\partial \mathcal{L}_{\text{cls}}}{\partial {\xinput_{1, 1}}} & \cdots & \frac{\partial \mathcal{L}_{\text{cls}}}{\partial {\xinput_{w,1}}} \\
  \vdots  & \ddots & \vdots  \\
  \frac{\partial \mathcal{L}_{\text{cls}}}{\partial {\xinput_{1,h}}} & \cdots & \frac{\partial \mathcal{L}_{\text{cls}}}{\partial {\xinput_{w,h}}} 
 \end{bmatrix}
\end{equation}

We can then compute the Jacobian's Fourier spectrum to retrieve a frequency profile of the model. Since the input images are made up of discrete pixels, we can extract this information by applying a discrete Fourier transform ($\mathcal{F}$) to the Jacobian to get a map ($M$) of its frequency components' magnitude:
\begin{equation}
    M_{i,j} = | \mathcal{F}(J)[i,j] | 
\end{equation}
In our experiments where the input images have 3 RGB channels, we compute the Fourier map for each other channel separately ($M_r$, $M_g$, $M_b$) and take the mean across these channels, i.e., $M_{i,j} = \frac{M_r[i,j] + M_g[i,j] + M_b[i,j]}{3} $.

Next, we propose to compute a scalar bias term ($\mathcal{B}_{\text{low}}$) from the 2-D map ($M$) to measure the relative bias (or ratio) of low-frequency components with respect to high-frequency components. One criterion for $\mathcal{B}_{\text{low}}$ is that the contribution of the frequency magnitude to this term should monotonically decrease as the frequency increase, i.e., larger high-frequency magnitudes results in lower $\mathcal{B}_{\text{low}}$ values. To satisfy this, we monotonically decrease the exponent value on the frequency magnitude as its frequency increases. For a 1-D scenario where $l$ is the dimension of the Fourier spectrum, we can express this bias term $\mathcal{B}_{\text{low}}$ as:
\begin{equation}
     \mathcal{B}_{\text{low}} = \Pi_i (M_i)^{\alpha_i},~~~~~ \alpha_i < \alpha_j,~~~~~ \forall i, j \in [1, l], ~~~~~  i < j
\end{equation}

where $M_1$ and $M_l$ are the magnitudes of the lowest and highest frequency components respectively. 
To ensure that $\mathcal{B}_{\text{low}}$ measures the relative ratio between the frequencies rather than absolute values of the frequency components, we use the following constraint on the $\alpha$ so that it is independent of the sum of the components' magnitudes:
\begin{equation}
     \alpha_i = - \alpha_{(l-i+1)}, ~~~~~ \forall i \in [1, l]
\end{equation}

In all our experiments, we use an array of values whose values are evenly spaced with distance $k$ from $\alpha_1$ and $\alpha_l$ for the $\alpha$ values, i.e,
\begin{equation}
     \alpha_{i+1} = \alpha_i + k, ~~ \forall i \in [1, l-1]
\end{equation}
and use $\alpha_1 = 1, \alpha_l = -1$.
When generalizing the bias term to two axes, we can compute the sum of all bias terms along each row and column of the 2-D Fourier spectrum to give:
\begin{equation}
    \mathcal{B}_{\text{low}} = \left[ \sum_j \left[ \Pi_i (M_{i,j})^{\alpha_i} \right] +  \sum_i \left[ \Pi_j (M_{i,j})^{\alpha_j}\right] \right] 
\end{equation}
In the next section, we discuss how $\mathcal{B}_{\text{low}}$ can be used to regularize neural networks' Jacobians to bias them towards low-frequency components.

\begin{figure}[!htbp]
    \centering
    \includegraphics[width=0.8\linewidth]{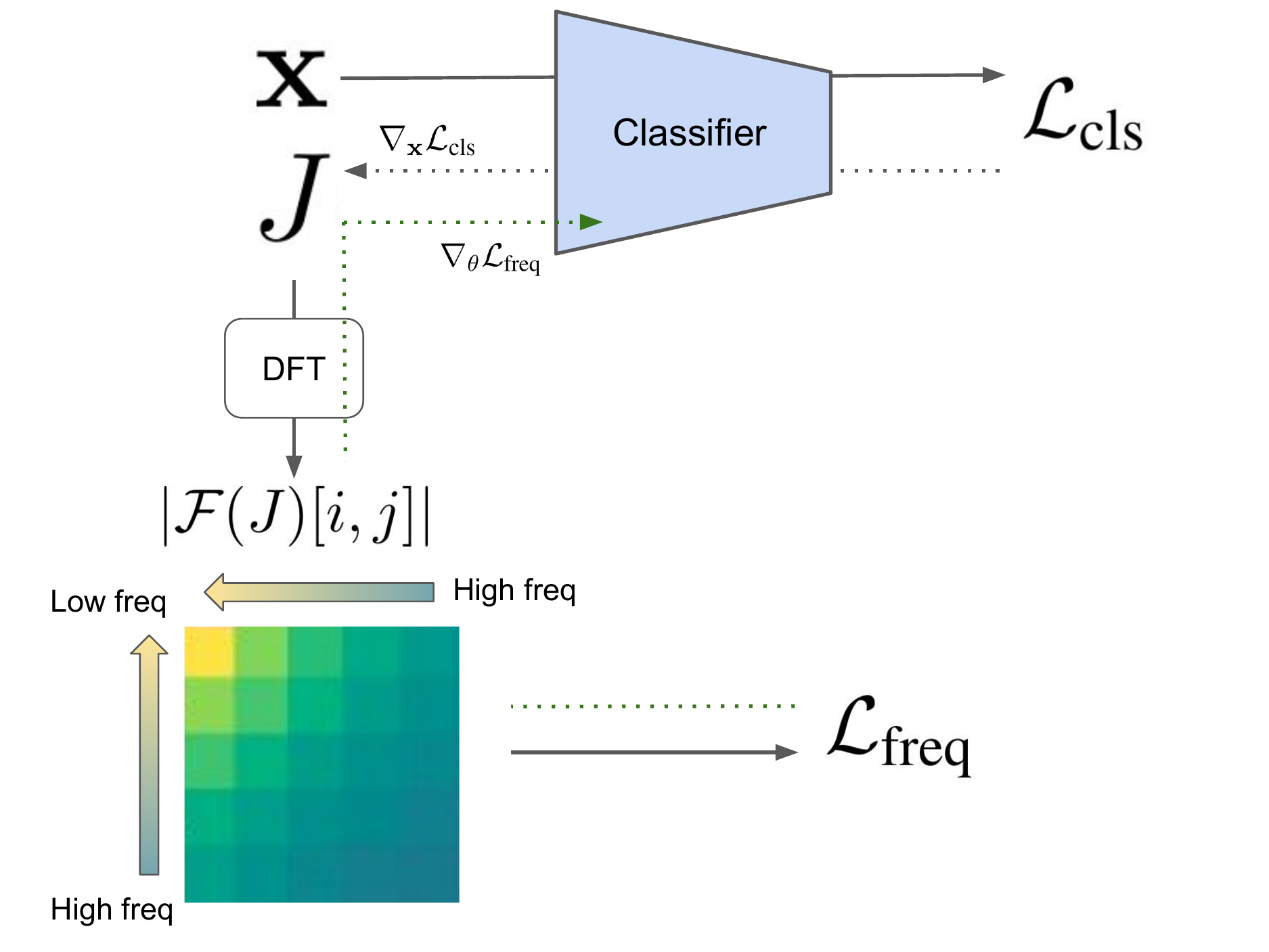}
    \caption{Training architecture of Jacobian frequency regularization (JaFR). JaFR trains the model's Jacobian to bias towards low-frequency components by shifting its Fourier spectrum's intensity towards the low-frequency regions, leftwards for the horizontal and upwards for the vertical axis.}
    \label{fig:JaFR}
\end{figure}

\subsection{Jacobian Frequency Regularization (JaFR)} \label{sec:JaFR training}
To recall, the aim here is to control the Fourier spectrum of a model's Jacobians to study how its Fourier profile can affect model robustness. To achieve this, we propose Jacobian Frequency Regularization (JaFR) to bias the Fourier spectrum of a model towards the low-frequency region. Figure~\ref{fig:JaFR} shows a summary of how our proposed Jacobian Frequency Regularization (JaFR) trains a model to alter its frequency profile.
Since the operations involved in computing $\mathcal{B}_{\text{low}}$ are differentiable, we can simply incorporate it in the following loss function with the aim to maximize the low-frequency bias of a model:

\begin{multline}
    \mathcal{L}_{\text{freq}}  = - \log \mathcal{B}_{\text{low}} = \\ - \log \left[ \sum_j \left[ \Pi_i (M_{i,j})^{\alpha} \right] +  \sum_i \left[ \Pi_j (M_{i,j})^{\beta}\right] \right]    
\end{multline}


Combining with the classification loss in Equation~\ref{eq:cls xent},  we can optimize through stochastic gradient descent to approximate the optimal parameters for the classifier $f_{\text{cls}}$ as follows,
\begin{equation}
\theta^* = \argmin_{\theta} (\mathcal{L}_{cls} + \lambda_{freq} \mathcal{L}_{freq})
\end{equation}
where $\lambda_{freq}$ determines the weight of JaFR in the model's training. A positive $\lambda_{freq}$ value regularizes the model to bias towards low-frequency features while a negative value (indicated as JaFR(-)) conversely steers the model towards high-frequency features. The $\mathcal{L}_{freq}$ term can be computed during each standard training iteration with an additional backpropagation step. Algorithm~\ref{algo:JaFR Training} summarizes the training of a classifier with JaFR. 
In the next sections, we present the results on the effect of JaFR on model robustness under adversarial examples and corruption noises.

\begin{algorithm}
\footnotesize
 \caption{Jacobian Frequency Regularization Training}
 \label{algo:JaFR Training}

\textbf{Input:} Train data $\mathcal{D}_{\text{train}}$, learning rate $\gamma$
\begin{algorithmic}[1]
\FOR{ each training iteration }
\STATE Sample $(\xinput, \mathbf{y}) \sim \mathcal{D}_{\text{train}}$
\STATE $\mathcal{L}_{\text{cls}} \gets - \mathbf{y}^\top \log f_{\text{cls}} (\xinput)$ \hfill\textit{(1) Compute classification cross-entropy loss}
\STATE $J \gets \nabla_{\xinput} \mathcal{L}_{\text{cls}}$ \hfill\textit{(2) Compute Jacobian matrix}
\STATE $M_{i,j} \gets | \mathcal{F}(J)[i,j] | $ \hfill\textit{(3) Compute frequency magnitudes}
\STATE $\mathcal{L}_{\text{freq}} \gets - \log \left[ \sum_j \left[ \Pi_i (M_{i,j})^{\alpha} \right] +  \sum_i \left[ \Pi_j (M_{i,j})^{\beta}\right] \right] $ \hfill\textit{(4) Compute frequency bias}
\STATE $\theta \gets \theta - \gamma ~ \nabla_{\theta} (\mathcal{L}_{\text{cls}} + \lambda_{\text{freq}} \mathcal{L}_{\text{freq}})  $ \hfill\textit{(5) Update the classifier $f_{\text{cls}}$ to minimize $\mathcal{L}_{cls}$ and $\mathcal{L}_{\text{freq}}$}
\ENDFOR
\end{algorithmic}
\end{algorithm}

\begin{table*}[!htbp]
    \centering
    \footnotesize
        \begin{tabular}{ l|cc|cccc|c }
         \hline
         ~ & Standard & JaFR & FGSM AT & FGSM AT & FGSM AT & PGD AT & $\mathcal{B}_{\text{low}}$ \\
         ~ & ~ & ~ & + JaFR(-) & ~ & + JaFR & ~ & ~ \\
        \hline
        Clean & 93.11$\pm$0.20 & 93.13$\pm$0.11 & 83.16 $\pm$1.01 & 84.80$\pm$1.37 & 79.94$\pm$0.22 & 79.31$\pm$0.23 & - \\
        \hline
        mCE & 100.00 & 95.84 & 127.51 & 124.32 & 134.37 & 135.71 & - \\
        \hline
        Fog & 86.23$\pm$0.45 & \textbf{86.60$\pm$0.28}  &   \textbf{67.15$\pm$3.81}   & 63.00$\pm$4.03  & 56.60$\pm$0.72 & 56.36$\pm$1.25 & 12.85\\
        Saturate & \textbf{89.27$\pm$0.16}  & 89.07$\pm$0.28 &  77.55$\pm$0.76   & \textbf{79.54$\pm$1.07} & 76.61$\pm$0.22 & 76.29$\pm$0.21 & 12.28\\
        Contrast & \textbf{75.10$\pm$1.08} & 73.45$\pm$0.95 &   \textbf{50.93$\pm$3.10}   & 44.06$\pm$3.97 & 41.96$\pm$0.61 & 41.19$\pm$0.88 & 12.14\\
        Bright & \textbf{91.56$\pm$0.15} & 91.45$\pm$0.18 &  81.02$\pm$0.88  & \textbf{83.04$\pm$1.20} & 76.95$\pm$0.42 & 76.01$\pm$0.58 & 12.01\\
        Snow & 79.54$\pm$0.71  & \textbf{81.06$\pm$0.15} &  75.05$\pm$1.52   & \textbf{77.83$\pm$1.43} & 74.28$\pm$0.27 & 73.73$\pm$0.21 & 11.53\\
        Frost & 75.41$\pm$0.63 & \textbf{78.97$\pm$0.56} &   75.82$\pm$1.64  & \textbf{77.70$\pm$1.44} & 70.50$\pm$0.54 & 69.06$\pm$1.18 & 10.61\\
        Motion & \textbf{75.67$\pm$1.35} & 75.13$\pm$1.21 &  67.33$\pm$2.01  & 65.15$\pm$3.47 & \textbf{72.65$\pm$0.41}  & 72.15$\pm$0.78 & 10.61\\
        Zoom & 76.42$\pm$1.65 & \textbf{76.48$\pm$1.20}  &   71.39$\pm$1.87  & 69.37$\pm$3.91 & \textbf{75.38$\pm$0.44} & 74.78$\pm$0.67 & 10.37\\
        Elastic & 81.62$\pm$0.60 & \textbf{81.98$\pm$0.55} &  73.88$\pm$1.45   & 73.90$\pm$2.27 & \textbf{74.86$\pm$0.37} & 74.25$\pm$0.58 & 10.05\\
        Pixel & \textbf{72.90$\pm$0.70} & 72.45$\pm$0.89 &   79.78$\pm$0.72   & \textbf{81.20$\pm$0.55} & 78.15$\pm$0.14 & 77.47$\pm$0.33 & 8.2\\
        Gauss. B & 73.38$\pm$1.58 & \textbf{73.53$\pm$1.16}  &  70.56$\pm$1.51   & 68.49$\pm$3.93 & \textbf{74.39$\pm$0.38} & 73.71$\pm$0.63 & 7.82\\
        Defocus & 81.57$\pm$0.83 & \textbf{81.76$\pm$0.71} &  74.22$\pm$1.37  & 73.42$\pm$3.01 & \textbf{76.15$\pm$0.29} & 75.46$\pm$0.54 & 7.49\\
        Glass & 50.18$\pm$1.75 & \textbf{55.12$\pm$0.58} &  66.28$\pm$2.89   & 70.18$\pm$1.07 & \textbf{73.75$\pm$0.28} & 73.33$\pm$0.49 & 7.01\\
        Spatter & 80.77$\pm$0.11 & \textbf{82.29$\pm$0.45} &  76.46$\pm$1.68  & \textbf{78.49$\pm$2.04} & 75.83$\pm$0.32 & 75.29$\pm$0.32 & 6.77\\
        JPEG & 77.76$\pm$0.56 & \textbf{79.56$\pm$0.60} &  80.38$\pm$0.77   & \textbf{81.47$\pm$1.54} & 78.19$\pm$0.18 & 77.58$\pm$0.24 & 6.51\\
        Speckle & 62.79$\pm$2.44 & \textbf{66.74$\pm$0.51} &  68.86$\pm$4.83  & 74.07$\pm$4.08 & \textbf{77.36$\pm$0.21} & 76.22$\pm$0.54 & 3.76\\
        Shot & 59.63$\pm$2.75 & \textbf{64.32$\pm$0.76} &  68.32$\pm$5.06   & 73.73$\pm$4.19 & \textbf{77.35$\pm$0.24} & 76.19$\pm$0.42 & 3.68\\
        Impulse  & 56.50$\pm$1.64 & \textbf{58.75$\pm$0.59} &   57.73$\pm$6.09  & 64.51$\pm$8.91 & 71.55$\pm$0.54 & \textbf{71.56$\pm$0.74} & 3.63\\
        Gauss. N & 48.70$\pm$3.31 & \textbf{54.34$\pm$1.00} &  62.91$\pm$6.74   & 69.40$\pm$5.27 & \textbf{76.59$\pm$0.33} & 75.15$\pm$0.49 & 3.62\\
        \hline
        \end{tabular}
    \caption{Accuracy values ($\uparrow$ better) and mCE ($\downarrow$ better) for different models under CIFAR-10 corruptions. The corruption types are arranged with descending order of low-frequency bias $\mathcal{B}_{\text{low}}$.}
\label{tab:cifar10 corruption}
\end{table*}

\section{Experiments}
We conduct experiments across 4 image datasets (SVHN, CIFAR-10 and CIFAR-100, TinyImageNet) \cite{netzer2011reading,krizhevsky2009learning} to study the effect of JaFR on model robustness against adversarial examples and common image corruptions.
We largely follow the training setting as \cite{andriushchenko2020understanding} where the PreAct ResNet-18 architecture \cite{he2016identity} is used for all models. To benchmark JaFR's effect on adversarial robustness, we compare with a relatively weak AT baseline (FGSM AT) and 2 strong AT baselines (FGSM AT + GradAlign \cite{andriushchenko2020understanding} and PGD AT \cite{madry2017towards}). All AT models use $\epsilon=\frac{8}{255}$. The dataset-specific parameters are further detailed in the following sections. Experiments are run on Nvidia V100 GPUs.

\textbf{SVHN}
All models were trained for 15 epochs, with a cyclic learning rate schedule and a max learning rate of 0.05. We use $\lambda_{freq}=0.001$ and $\lambda_{freq}=0.05$ for the JaFR and FGSM AT + JaFR model respectively. FGSM + GradAlign model uses the same hyperparameters in \cite{andriushchenko2020understanding} for training.

\textbf{CIFAR-10}
The training setup largely follows the SVHN experiments where the PreAct ResNet-18 architecture is used for all models. All models were trained for 30 epochs, with a cyclic learning schedule. The standard-trained, JaFR and FGSM models use a max learning rate of 0.2 while the FGSM + JaFR, FGSM + GradAlign and PGD AT models used a max learning rate of 0.3. Both JaFR and FGSM AT + JaFR models use $\lambda_{freq}=0.001$. 

\textbf{CIFAR-100}
The training setup is similar to that in CIFAR-10, except for the learning rates and $\lambda_{freq}=0.001$ values. The max learning rates for all models are 0.3 except for the JaFR (0.1) and standard-trained (0.2) models.

\textbf{TinyImageNet}
The training setup is similar to that in CIFAR-100, except for the $\lambda_{freq}=0.002$ value.

\subsection{Common Corruption Robustness}
CIFAR-10-C and CIFAR-100-C are common corruption benchmarks where the CIFAR-10/CIFAR-100 test sets are corrupted with an array of 19 corruption types with 5 levels of severity. The accuracy values are reported as the average over the 5 severities of each corruption. Table~\ref{tab:cifar10 corruption stats} (in Appendix) shows top-6 corruptions that have the highest ratio of low- and high-frequency components as measured by $\mathcal{B}_{\text{low}}$.


\textbf{JaFR improves corruption robustness for standard-trained models.} From Table~\ref{tab:cifar10 corruption} and \ref{tab:cifar100 corruption} (Appendix), we observe that the JaFR model achieves the best overall corruption robustness for both CIFAR-10 and -100 images, as indicated by its lowest relative mean corruption errors (mCE). The JaFR outperforms the second-best model (standard-trained) in 14 out of 19 corruptions for CIFAR-10-C and 17 out of 19 corruptions for CIFAR-100-C test samples. This is an encouraging sign for direct frequency regularization of neural networks as an approach for better corruption robustness. 

When comparing the spectra of the standard and JaFR model in Table~\ref{tab:cifar10 model fourier profile}, we observe that the intensities shift from the highest-frequency (bottom-right) region to lower frequency regions (left and top) with JaFR. This marks a lower reliance on high-frequency features in the model, which can confer better robustness against high-frequency corruptions. Indeed, with JaFR, we see the biggest improvements for high-frequency corruptions (e.g., Glass, Shot and Gaussian noise). Moreover, we can see that the spectra of our JaFR-only model, unlike the adversarially trained models (e.g., PGD AT), do not concentrate the intensities on the lowest-frequency region (top-left) which may result in an over-reliance on low-frequency features and make the model more susceptible to low-frequency corruptions. During our experiments, larger $\lambda_{freq}$ values for JaFR do concentrate the intensities even more in the low-frequency regions but result in poorer overall performance. This balance in reliance on both high and low-frequency features may explain the improved overall performance of JaFR against corruption.

\begin{table*}[!htbp]
    \centering
    \footnotesize
    \setlength{\tabcolsep}{0.4em}
        \begin{tabular}{ l|cccccccc }
         \hline
         Model & Standard & JaFR & FGSM AT & FGSM AT & FGSM AT & FGSM AT & PGD AT & Original \\
         ~ & ~ & ~ & ~ & + JaFR(-) & + JaFR & + GradAlign & ~ & Image \\
         \hline
         $\mathbb{E} \left[| \mathcal{F}(J) | \right]$ & 
         \parbox[c]{4em}{\includegraphics[width=4em]{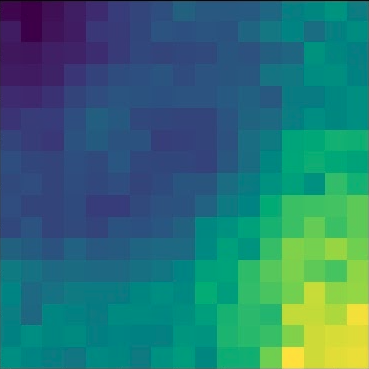}} & 
         \parbox[c]{4em}{\includegraphics[width=4em]{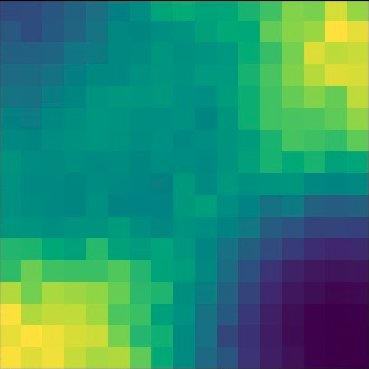}} & 
         \parbox[c]{4em}{\includegraphics[width=4em]{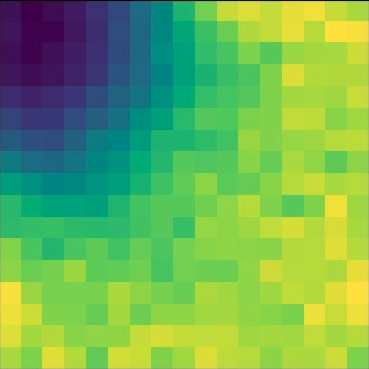}} & 
         \parbox[c]{4em}{\includegraphics[width=4em]{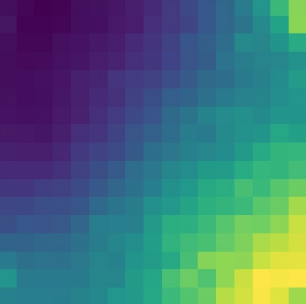}} & 
         \parbox[c]{4em}{\includegraphics[width=4em]{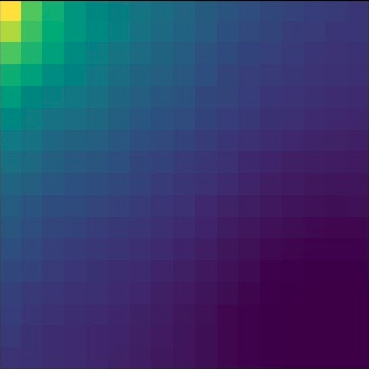}} & 
         \parbox[c]{4em}{\includegraphics[width=4em]{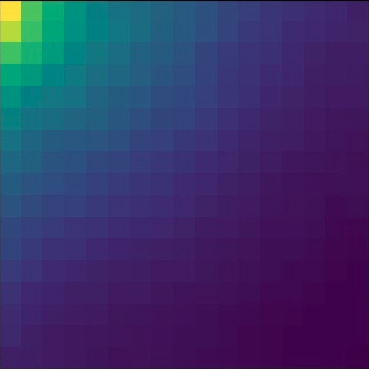}} & 
         \parbox[c]{4em}{\includegraphics[width=4em]{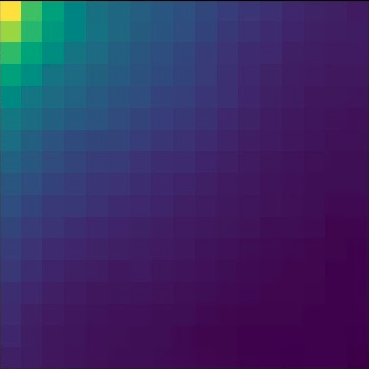}} & - \\
         \vspace{0.02cm} \\
         $ J $ &          
         \parbox[c]{4em}{\includegraphics[width=4em]{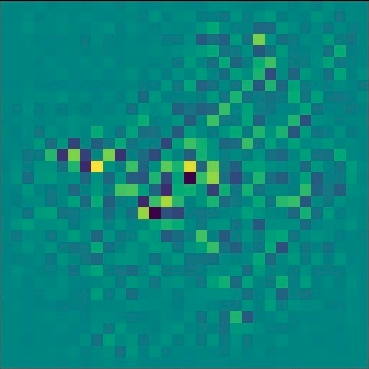}} & 
         \parbox[c]{4em}{\includegraphics[width=4em]{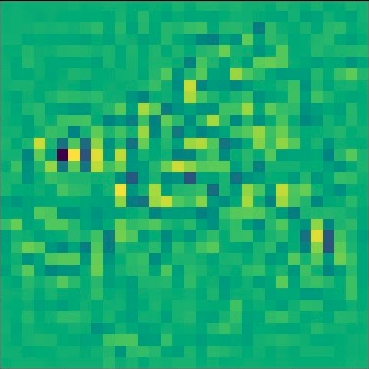}} & 
         \parbox[c]{4em}{\includegraphics[width=4em]{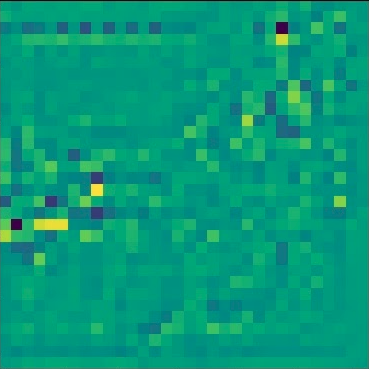}} & 
         \parbox[c]{4em}{\includegraphics[width=4em]{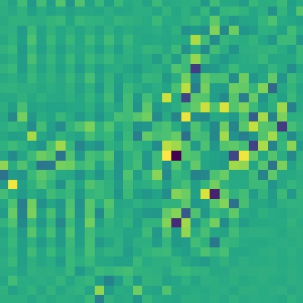}} & 
         \parbox[c]{4em}{\includegraphics[width=4em]{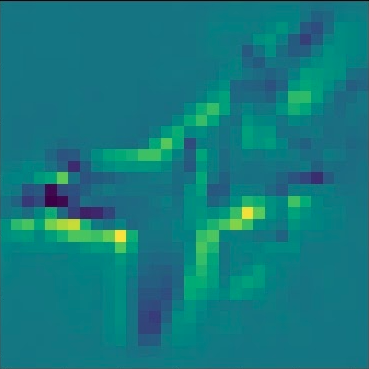}} & 
         \parbox[c]{4em}{\includegraphics[width=4em]{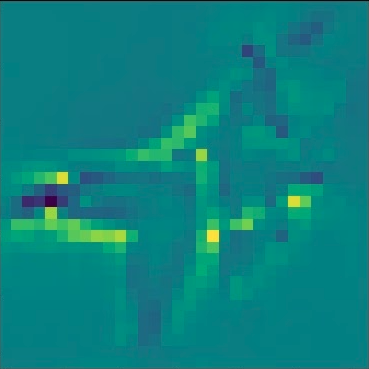}} & 
         \parbox[c]{4em}{\includegraphics[width=4em]{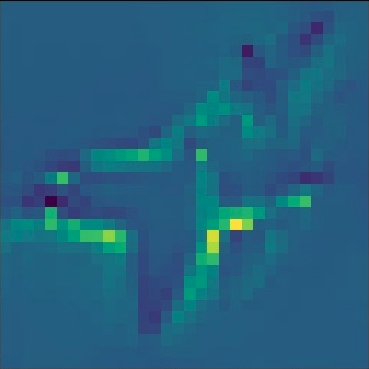}} & 
         \parbox[c]{4em}{\includegraphics[width=4em]{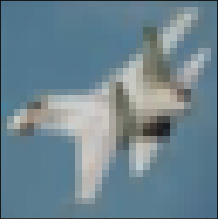}} \\
         \hline
         $\mathcal{B}_{\text{low}}$ & 1.56 & 16.71 & 2.68 & 0.326 & 15.41 & 7.88 & 7.49 & -\\
        \end{tabular}
        
    \caption{CIFAR-10 models' low-frequency bias ($\mathcal{B}_{\text{low}}$), frequency spectra and Jacobians.}
\label{tab:cifar10 model fourier profile}
\end{table*}

\textbf{Tradeoff between robustness against low- and high-frequency corruption exists.} For both CIFAR-10-C (Table~\ref{tab:cifar10 corruption}) and CIFAR-100-C (Table~\ref{tab:cifar100 corruption}), we observe that when JaFR is added to FGSM AT, the performance against low-frequency corruptions (e.g., Fog \& Contrast) drops when the accuracy against high-frequency corruptions (e.g., Impulse \& Gaussian noise) improves. We also see such behavior when JaFR is added to the standard trained model in CIFAR-10-C. This aligns with previous studies that show such a tradeoff between corruptions with different frequency profiles. 

To study the opposite effect of JaFR, a JaFR(-) model variant is added where the $\lambda_{freq}$ term has a negative value to investigate the effect of biasing the model towards high-frequency components.
We can see such tradeoff emerging with an opposite effect when JaFR(-) is combined with FGSM AT to bias the model towards high frequency features, where the performance against high-frequency corruptions (e.g., Impulse \& Gaussian noise) drops when the accuracy against low-frequency corruptions (e.g., Fog \& Contrast) improves. This suggests that direct frequency regularization of models is a promising way to tailoring robustness against a set of corruptions that is more relevant for their applications.

\textbf{FGSM AT + JaFR outperforms PGD AT in almost all corruptions.} From our experiments, apart from one case of Impulse corruption in CIFAR-10-C, the FGSM AT + JaFR outperforms the PGD AT model in all the other 37 corruptions scenarios. We speculate that the PGD AT model might have overfitted to resist adversarial perturbations, making them excessively reliant on a small subset of low-frequency features that can be easily disrupted by the common corruptions.

\subsection{Adversarial Robustness}
We evaluate the adversarial robustness of models with FGSM \cite{goodfellow2014explaining} and PGD \cite{madry2017towards} attacks. PGD uses 50 gradient iterations and 10 restarts with a step size of $\alpha_{adv}=\epsilon/4$.

\textbf{Biasing towards low frequency can boost adversarial robustness in weakly adversarially trained models.}
For all the four image datasets, when JaFR is combined with FGSM, there is a significant boost in both clean and adversarial accuracy values from the FGSM AT model for all attack types (see Table~\ref{tab:svhn}, \ref{tab:cifar10}, \ref{tab:cifar100} and \ref{tab:tinyimagenet}). This observation is similar to what is observed for a recent defense, GradAlign, which also requires combining with FGSM to improve the adversarial robustness of the model. The need of using FGSM training samples to see an improvement in JaFR's adversarial robustness indicates that strong adversarial examples are more than just high-frequency noise and are able to find regions of high error in the uneven loss landscape of a non-adversarially trained model \cite{liu2020loss}. For the CIFAR-10 and -100, the improvement of adding JaFR to FGSM is large enough to achieve a robustness level that is competitive with the strong PGD AT baseline. In Table C, we observe that JaFR scales well to a larger dataset, with a smaller amount of computation than PGD AT. Conversely, combining with JaFR(-) to bias the FGSM AT model towards high frequency with a negative $\lambda_{freq}$ term was observed in our experiments to worsen its adversarial robustness against PGD attacks. Experiments (Table~\ref{tab:more attacks} in Appendix) on more advanced attacks such as CW \cite{carlini2017towards} and AutoAttack \cite{croce2020reliable} show that the robustness gain from JaFR can resist even stronger attacks and is not due to gradient masking. Moreover, from Table~\ref{tab:tinyimagenet}, we observe that JaFR scales well, with a smaller amount of computation than the PGD AT. 

\begin{table}[!htbp]
    \centering
    \footnotesize
        \begin{tabular}{ l|ccc }
         \hline
         Model & Clean & FGSM & PGD \\
         \hline
        Standard & 96.62$\pm$0.05 &  21.86$\pm$0.90 & 0.17$\pm$0.02 \\
        JaFR &  96.58$\pm$0.07  &  21.94$\pm$1.16 & 0.14$\pm$0.03  \\
        FGSM AT &  92.33$\pm$0.20  &  89.16$\pm$5.87 & 0.52$\pm$0.90   \\
        FGSM + JaFR & 87.22$\pm$4.63 & 59.4$\pm$12.75 & 19.64$\pm$16.26   \\
        FGSM + GA & 92.51$\pm$0.26 & 59.52$\pm$0.26 & 46.63$\pm$0.15 \\
        PGD AT & 92.13$\pm$0.51 & 62.38$\pm$0.86 & 56.79$\pm$0.29 \\
         \hline
        \end{tabular}
    \caption{SVHN accuracy (\%) on clean and adversarial test samples.}
\label{tab:svhn}
\end{table}

\begin{table}[!htbp]
    \centering
    \footnotesize
        \begin{tabular}{ l|ccc }
         \hline
         Model & Clean & FGSM & PGD \\
         \hline
        Standard & 93.11$\pm$0.20 & 16.04$\pm$0.85 & 0 \\
        JaFR & 93.13$\pm$0.11 & 15.86$\pm$0.88 & 0 \\
        FGSM AT & 84.80$\pm$1.37 & 85.14$\pm$4.01 & 0.01$\pm$0.02 \\
        FGSM + JaFR & 79.94$\pm$0.22 & 53.12$\pm$0.25 & 46.32$\pm$0.15 \\
        FGSM + GA & 80.07$\pm$0.21 & 54.26$\pm$0.55 & 46.97$\pm$0.15 \\
        PGD AT & 79.31$\pm$0.23 & 54.00$\pm$0.72 & 49.63$\pm$0.20 \\
         \hline
        \end{tabular}
    \caption{CIFAR-10 accuracy (\%) on clean and adversarial test samples.}
\label{tab:cifar10}
\end{table}

\textbf{JaFR removes high-frequency components in models' Jacobian.} From Table~\ref{tab:cifar10 model fourier profile}, we can see that models with JaFR have the highest low-frequency bias ($\mathcal{B}_{\text{low}}$) values, indicating that the training regularization is successful in increasing the low-frequency components of the model's Jacobians. Furthermore, from the Jacobian's Fourier spectra ($\mathbb{E} \left[| \mathcal{F}(J) | \right]$), intensities of the spectra shift from the high-frequency regions (right and bottom) towards the left and top regions of the spectra which indicate the low-frequency components along the horizontal and vertical axis respectively. In contrast, using a negative $\lambda_{freq}$ term (in FGSM AT + JaFR(-)) concentrates the spectra towards highest-frequency (bottom-right) region and drastically lowers the ($\mathcal{B}_{\text{low}}$) value which indicates a strong reliance in high-frequency features.

\begin{table}[!htbp]
    \centering
    \footnotesize
        \begin{tabular}{ l|ccc }
         \hline
         Model & Clean & FGSM & PGD \\
         \hline
        Standard & 72.27$\pm$0.31 & 8.04$\pm$0.84 & 0.13$\pm$0.03 \\
        JaFR & 74.59$\pm$0.37 & 8.60$\pm$0.81 & 0.08$\pm$0.04 \\
        FGSM AT & 52.08$\pm$2.85 & 31.43$\pm$2.15 & 0.04$\pm$0.04\\
        FGSM + JaFR & 50.49$\pm$0.21 & 26.96$\pm$0.36 & 23.42$\pm$0.27 \\
        FGSM + GA & 50.68$\pm$0.28 & 27.44$\pm$0.68 & 23.93$\pm$0.19  \\
        PGD AT & 49.57$\pm$0.44 & 27.68$\pm$0.77 & 25.24$\pm$0.25 \\
         \hline
        \end{tabular}
    \caption{CIFAR-100 accuracy (\%) on clean and adversarial test samples.}
\label{tab:cifar100}
\end{table}

\begin{table}[!htbp]
    \centering
    \footnotesize
        \begin{tabular}{ l|ccc|c }
         \hline
         Model & Clean & FGSM & PGD & Complex. (s/iter) \\
         \hline
        Standard & 63.59 & 2.23 & 0.05 & 0.1575 \\
        JaFR & 63.44 & 1.87 & 0.04 & 0.6015  \\
        FGSM AT & 45.89 & 17.54 & 0 & 0.4245 \\
        FGSM + JaFR &  47.43 & 20.3 & 18.86 & 0.6027  \\
        FGSM + GA & 40.69 & 19.86 & 17.81 & 0.6003  \\
        PGD AT & 40 & 17.54 & 19.91 & 1.34 \\
         \hline
        \end{tabular}
    \caption{Model accuracy and computational complexity on the TinyImageNet dataset.}
\label{tab:tinyimagenet}
\end{table}

\textbf{JaFR improves saliency of Jacobians.} When JaFR is added to the FGSM AT model, we observe that its Jacobians ($J$) become more salient (see Table~\ref{tab:cifar10 model fourier profile}). Together with the boost in adversarial robustness from the FGSM AT model, this corroborates previous results that saliency of Jacobian is correlated with adversarial robustness \cite{etmann2019connection}.

\section{Conclusion}
Model robustness is growing more important as deep learning models are gaining wider adoption. Here, we delve further into the link between the frequency characteristic of a model and its robustness by proposing Jacobian frequency bias. Through this term, we can control the distribution of high- and low-frequency components in the model's Jacobian and find that it can affect both corruption and adversarial robustness. We hope that our findings here will open an avenue for future work to explore other frequency-focused approaches to improve model robustness.


\bibliographystyle{named}
\bibliography{ijcai22}

\clearpage

\appendix

\section{Appendix}

\subsection{Background and Related Work}
\textbf{Adversarial Training} One of the most effective approaches to train models robust against adversarial examples is adversarial training (AT) \cite{goodfellow2014explaining}. The intuition behind AT is to match the training distribution with the adversarial example distribution. This is achieved by crafting adversarial examples in each training iteration and using them as training samples to minimize the following loss:

\begin{equation} \label{eq:AT objective appendix}
\mathcal{L}(\xinput, \mathbf{y}) = \mathbb{E}_{(\xinput, \mathbf{y}) \sim D} \left[ \max_{\delta \in B(\varepsilon)} \mathcal{L}(\xinput + \delta, \mathbf{y}) \right]
\end{equation}
where the inner maximization, $\max_{\delta \in B(\varepsilon)} \mathcal{L}(\xinput + \delta, \mathbf{y})$, is performed with gradient descent. The earliest form of AT, fast gradient sign method (FGSM) \cite{goodfellow2014explaining}, uses one single gradient to create adversarial training samples but models trained this way have shown to be vulnerable to stronger adversarial examples that are crafted with more gradient steps. The more recent projected gradient descent (PGD) adversarial training \cite{madry2017towards} and its variants \cite{zhang2019defense,qin2019adversarial,andriushchenko2020understanding,wu2020adversarial} are one of the stronger defenses which performs the following gradient step iteratively:
\begin{equation} \label{eq:pgd step}
    \delta \gets \mathrm{Proj} \left[ \delta - \eta ~ \text{sign} \left( \nabla_{\delta} \mathcal{L}(\xinput + \delta, \mathbf{y}) \right) \right ]
\end{equation}
where $\mathrm{Proj}(\xinput) = \argmin_{\zeta \in B(\varepsilon)} \| \xinput - \zeta \|$. Different from the AT-based techniques, JaFR does not rely on adversarial examples for training and is orthogonal to these AT-based techniques.

\textbf{Link between Frequency and Robustness} \label{sec:fourier perspectives on robustness appendix}
Studies have shown that data augmentation techniques have improved robustness against certain corruptions but degrade performances on others \cite{yin2019fourier}. \cite{yin2019fourier} carried out frequency analysis which shows a connection between the Fourier spectrum of a particular corruption type and whether its performance improves or degrades. More specifically, they found that Gaussian data augmentation and adversarial training result in models that rely heavily on low-frequency features in images. These models are hence more resistant against high-frequency noise but more susceptible to low-frequency corruptions. The authors analyzed the Fourier profile of trained models by measuring performance when noise is created by components bounded within a region of the Fourier spectrum or when training data have certain frequency components removed. \cite{yin2019fourier} use a qualitative inspection of the Fourier spectrum to show which corruption types have a relative ratio of low or high-frequency components while we propose a frequency bias term to quantify it. In other works that explored the link between frequency and robustness, \cite{sharma2019effectiveness} showed that models trained against adversarial examples are equally vulnerable as standard-trained models when faced with adversarial perturbations that are constrained to the low-frequency regions, confirming the findings from \cite{yin2019fourier} that adversarially trained models' robustness is based on invariance to high-frequency signals. \cite{ortiz2020hold} showed that trained models can learn to be invariant to low or high-frequency features depending on how discriminative these features are to the classification label (i.e., how much they change the label). \cite{tsuzuku2019structural} found that universal adversarial perturbations that fool a range of CNN classifiers are a combination of a few Fourier basis functions. \cite{vasconcelos2021impact} studied the effect of aliasing on models' performance. 
Our work here takes the Fourier analysis in a different direction by studying models through the Fourier spectrum of their Jacobians rather than their test or training data. More concretely, with the original training data, we train models and bias the frequency profile of the model's Jacobians towards low-frequency regions to see its effect on model robustness. 

\textbf{Jacobians of Robust Models} \label{sec:jacobians of robust models appendix}
The Jacobian, 
\begin{equation}
    J \coloneqq \nabla_{\xinput} \mathcal{L(\xinput, \mathbf{y})} 
\end{equation}
defines how the model's prediction changes with an infinitesimally small change to the input $\xinput$. For image classification, Jacobians can be loosely interpreted as a map of which pixels affect the model's prediction the most and, hence, give an illustration of important regions in an input image \cite{smilkov2017smoothgrad,adebayo2018sanity,ilyas2019adversarial}. Previous studies have observed that adversarially robustness models trained with AT display Jacobians that are more salient than those from non-robust standard-trained models \cite{tsipras2018robustness}. Since then, there a line of work that studies the theoretical link between the saliency of the Jacobians and robustness \cite{etmann2019connection} and exploits this link to improve robustness by regularizing for Jacobians' saliency \cite{chan2019jacobian,chan2020thinks}. By using generative adversarial networks (GANs) to train a model's Jacobians to fit the distribution of either the input images \cite{chan2019jacobian} or of a robust teacher model \cite{chan2020thinks}, adversarial robustness can be improved. In contrast, the regularizing effect of JaFR acts directly on the Fourier spectrum of the Jacobians rather than fitting them to a target distribution through GANs. Furthermore, our work here also studies the effect of JaFR on common corruptions, on top of adversarial robustness.

Other work that regularizes the input gradients to boost adversarial robustness includes using double backpropagation \cite{drucker1991double} to minimize the input gradients' Frobenius norm \cite{ross2018improving,jakubovitz2018improving}. Those approaches aim to constrain the effect that changes at individual pixels have on the classifier's output but not the frequency profile of neural networks like our method. Rather than aiming to improve the adversarial robustness, the core aim of our paper here is to investigate the relationship between the Fourier profile of models and robustness against corruptions.

\begin{table}[!htbp]
    \centering
    \footnotesize
    \setlength{\tabcolsep}{3pt}
        \begin{tabular}{ l|cccc }
         \hline
         Attack & FGSM AT & FGSM+JaFR & FGSM+GA & PGD AT \\
         \hline
        CW & 0 & 45.36 & 46.28 & 48.06  \\
        AutoAttack & 0 & 43.19 & 43.74 & 46.58 \\
         \hline
        \end{tabular}
    \caption{CIFAR-10 accuracy (\%) on advanced attacks.}
\label{tab:more attacks}
\end{table}


\begin{table*}[!htbp]
    \centering
    \footnotesize
        \begin{tabular}{ l|cc|cccc|c }
         \hline
         ~ & Standard & JaFR & FGSM AT & FGSM AT & FGSM AT & PGD AT & $\mathcal{B}_{\text{low}}$ \\
         ~ & ~ & ~ & + JaFR(-) & ~ & + JaFR & ~ & ~ \\
        \hline
        Clean & 72.27$\pm$0.31 & 74.59$\pm$0.37 & 53.31$\pm$1.60 & 52.08$\pm$2.85 & 50.49$\pm$0.21 & 49.57$\pm$0.44 & - \\
        \hline
        mCE & 100.00 & 96.29 &    117.04     & 120.03 & 114.70 & 116.50 & - \\
        \hline
        Fog & 58.58$\pm$0.54 & \textbf{61.97$\pm$0.52} &    \textbf{32.09$\pm$1.23}    & 31.83$\pm$3.36 & 27.39$\pm$0.45  & 26.69$\pm$0.78 & 12.91\\
        Saturate & 59.48$\pm$0.26 & \textbf{61.11$\pm$0.24} &    \textbf{41.05$\pm$1.56}     & 39.98$\pm$2.12 & 38.93$\pm$0.09 &  38.53$\pm$0.45 & 12.33\\
        Contrast & 48.88$\pm$0.72 & \textbf{51.78$\pm$0.66} &    \textbf{22.42$\pm$0.66}     & 22.27$\pm$2.84 & 20.22$\pm$0.41 & 19.38$\pm$0.74 & 12.15\\
        Bright & 67.69$\pm$0.20 & \textbf{70.19$\pm$0.34} &    \textbf{49.19$\pm$2.11}     & 48.63$\pm$2.58 & 44.87$\pm$0.31 & 44.28$\pm$0.49 & 11.98\\
        Snow & 50.46$\pm$0.28 & \textbf{54.50$\pm$0.49} &    \textbf{44.05$\pm$1.98}     &  41.96$\pm$2.81 & 42.44$\pm$0.49 &  42.25$\pm$0.78 & 11.56\\
        Frost & 45.82$\pm$0.95 & \textbf{52.18$\pm$0.47} &    \textbf{42.61$\pm$1.23}     & 42.06$\pm$2.45 & 37.22$\pm$0.65 & 36.54$\pm$1.07 & 10.77\\
        Motion & 51.22$\pm$0.56 & \textbf{52.22$\pm$0.96} &    37.90$\pm$2.28     & 36.73$\pm$4.48 & \textbf{43.24$\pm$0.36} & 42.16$\pm$0.63 & 10.63\\
        Zoom & 50.44$\pm$1.18 & \textbf{51.25$\pm$1.54} &    40.36$\pm$2.39     & 39.13$\pm$4.59 & \textbf{45.40$\pm$0.25} & 44.19$\pm$0.71 & 10.34\\
        Elastic & 56.66$\pm$0.31 & \textbf{58.45$\pm$0.48} &    43.88$\pm$1.45     & 42.17$\pm$3.44 & \textbf{44.88$\pm$0.24} & 43.93$\pm$0.47 & 10.03\\
        Pixel & 48.19$\pm$1.08 & \textbf{50.41$\pm$0.49} &    50.54$\pm$1.15     & 48.82$\pm$2.31 & \textbf{48.89$\pm$0.24} & 47.86$\pm$0.48 & 8.11\\
        Gauss. B & 48.64$\pm$1.22 & \textbf{48.78$\pm$1.49} &    39.71$\pm$2.17     & 38.42$\pm$4.46 & \textbf{44.43$\pm$0.24 }& 43.31$\pm$0.59 & 7.66\\
        Defocus & 56.52$\pm$0.89 & \textbf{57.44$\pm$1.22} &    43.44$\pm$1.65     & 42.12$\pm$3.96 & \textbf{46.15$\pm$0.21} & 44.99$\pm$0.49 & 7.38\\
        Glass & 21.32$\pm$0.37 & \textbf{24.79$\pm$0.46} &    41.14$\pm$3.26     & 38.01$\pm$3.00 & \textbf{45.24$\pm$0.24} & 44.33$\pm$0.40 & 7.02\\
        Spatter & 54.94$\pm$0.56 & \textbf{57.21$\pm$0.88} &    46.04$\pm$1.82     & 44.91$\pm$2.89 & \textbf{45.70$\pm$0.30} & 44.91$\pm$0.39 & 6.76\\
        JPEG & \textbf{52.04$\pm$0.43} & 50.40$\pm$0.65 &    50.17$\pm$1.31     & 48.33$\pm$2.79 & \textbf{48.50$\pm$0.33} & 47.48$\pm$0.43 & 6.6\\
        Speckle & 32.23$\pm$0.79 & \textbf{34.56$\pm$0.58} &    38.44$\pm$4.40     & 34.58$\pm$4.34 & \textbf{46.26$\pm$0.43} & 44.46$\pm$0.68 & 3.86\\
        Shot & 31.49$\pm$0.84 & \textbf{33.46$\pm$0.61} &    37.86$\pm$4.79     & 33.95$\pm$4.76 & \textbf{47.09$\pm$0.37} & 45.43$\pm$0.57 & 3.74\\
        Gauss. N& 24.21$\pm$0.78 & \textbf{25.14$\pm$0.56} &    32.47$\pm$6.02     & 28.47$\pm$6.33 & \textbf{46.32$\pm$0.44} & 44.57$\pm$0.49 & 3.68\\
        Impulse & \textbf{26.52$\pm$1.54} & 24.20$\pm$2.07 &    28.57$\pm$4.89     & 27.31$\pm$6.27 & \textbf{37.40$\pm$0.87} & 37.10$\pm$0.94 & 3.64\\
        \hline
        \end{tabular}
    \caption{Accuracy values ($\uparrow$ better) and mCE ($\downarrow$ better) for different models under CIFAR-100 corruptions. The corruption types are arranged with descending order of low-frequency bias $\mathcal{B}_{\text{low}}$.}
\label{tab:cifar100 corruption}
\end{table*}

\begin{table*}[!htbp]
    \centering
    \footnotesize

        \begin{tabular}{ l|cccccc }
         \hline
         Corruption & fog & saturate & contrast & brightness & snow & frost \\
         \hline
         $\mathcal{B}_{\text{low}}$ & 12.85 & 12.28 & 12.14 & 12.01 & 11.53 & 10.61  \\
         $\mathbb{E} \left[| \mathcal{F}(C(\xinput) - \xinput) | \right]$ & 
         \parbox[c]{4em}{\includegraphics[width=4em]{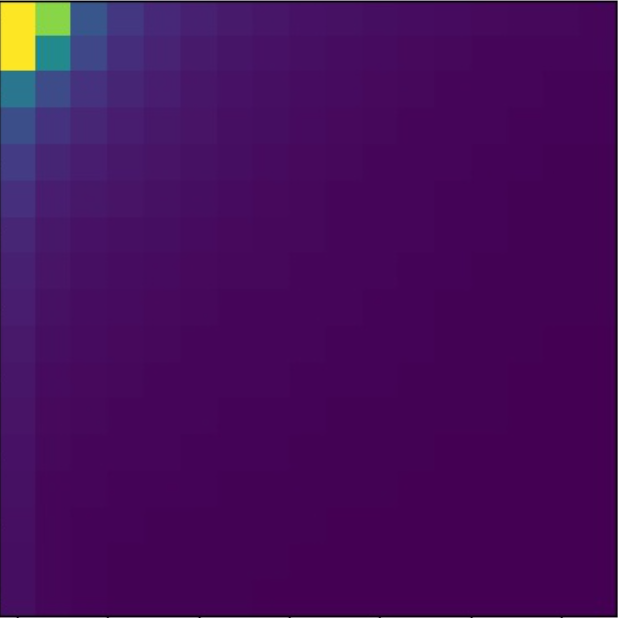}} & 
         \parbox[c]{4em}{\includegraphics[width=4em]{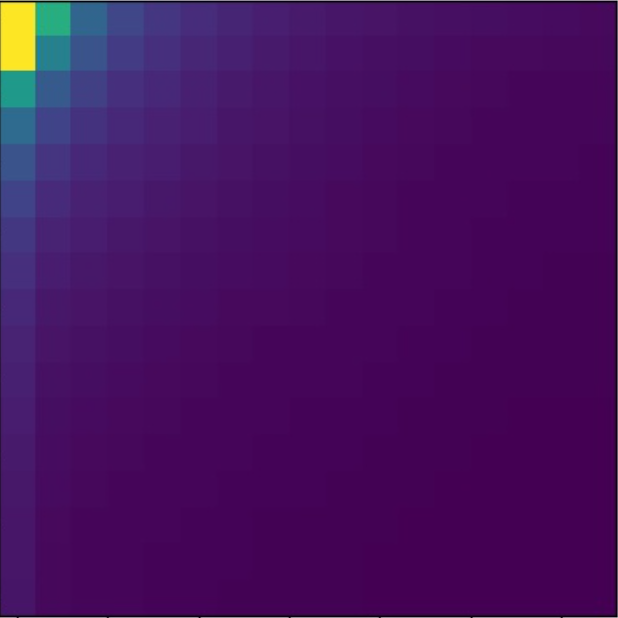}} & 
         \parbox[c]{4em}{\includegraphics[width=4em]{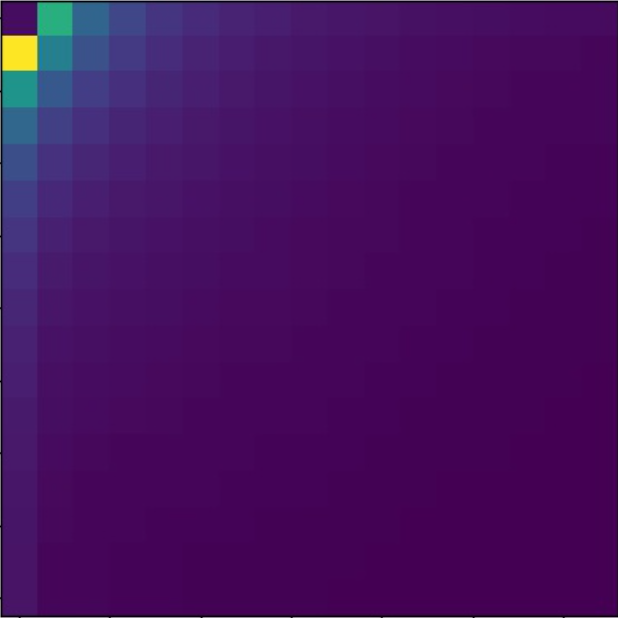}} & 
         \parbox[c]{4em}{\includegraphics[width=4em]{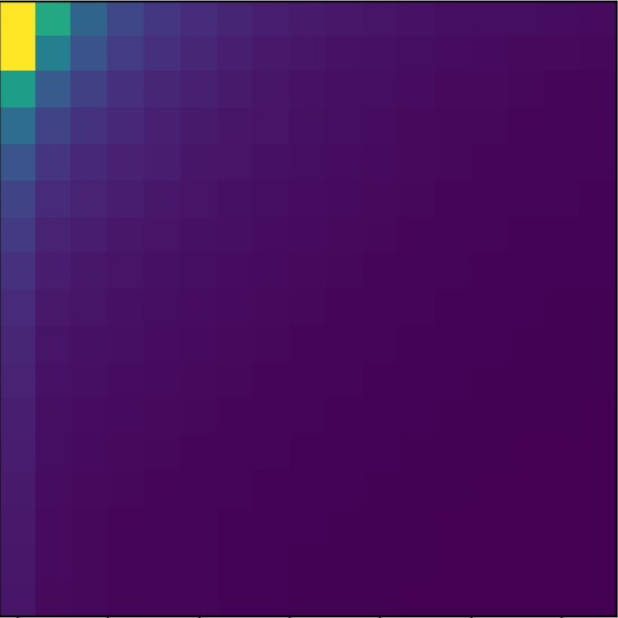}} & 
         \parbox[c]{4em}{\includegraphics[width=4em]{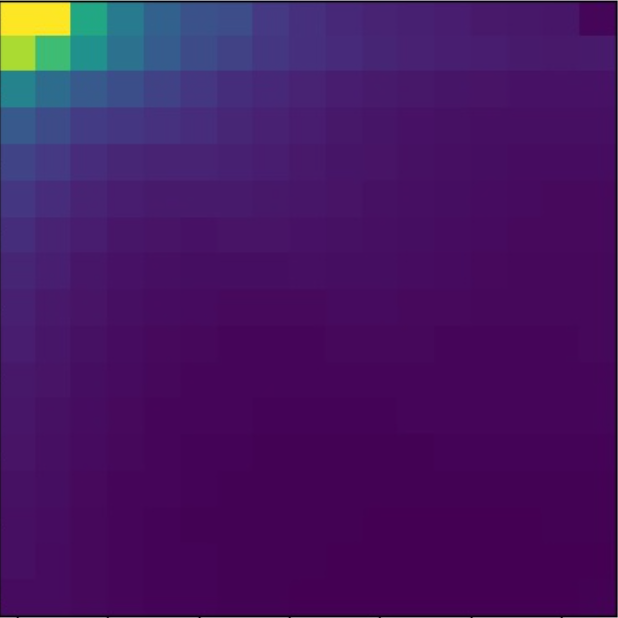}} & 
         \parbox[c]{4em}{\includegraphics[width=4em]{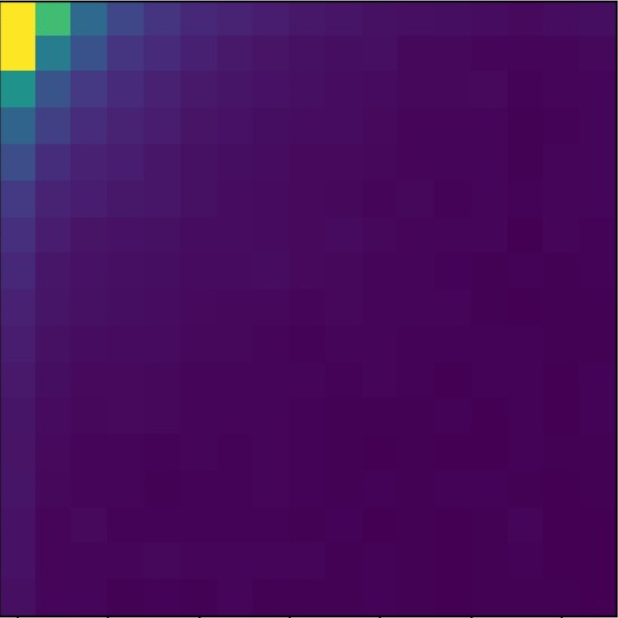}} \\
         \vspace{0.01cm} \\
         $ C(\xinput) $ &          
         \parbox[c]{4em}{\includegraphics[width=4em]{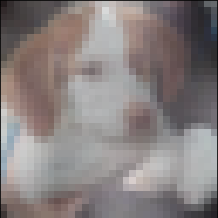}} & 
         \parbox[c]{4em}{\includegraphics[width=4em]{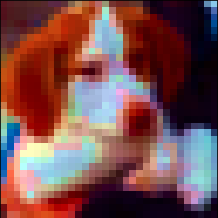}} & 
         \parbox[c]{4em}{\includegraphics[width=4em]{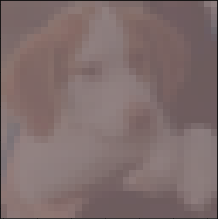}} & 
         \parbox[c]{4em}{\includegraphics[width=4em]{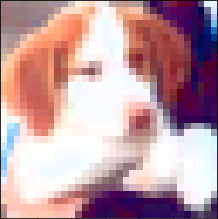}} & 
         \parbox[c]{4em}{\includegraphics[width=4em]{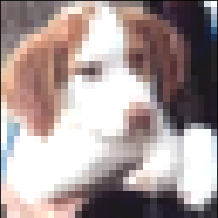}} & 
         \parbox[c]{4em}{\includegraphics[width=4em]{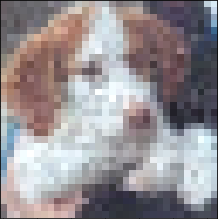}} \\
        \end{tabular}
        
        \vspace{1em}
        
        \begin{tabular}{ l|cccccc }
         \hline
         Corruption &  spatter & jpeg & speckle & shot & impulse & gauss noise \\
         \hline
         $\mathcal{B}_{\text{low}}$ & 6.77 & 6.51 & 3.76 & 3.68 & 3.63 & 3.62 \\
         $\mathbb{E} \left[| \mathcal{F}(C(\xinput) - \xinput) | \right]$ & 
         \parbox[c]{4em}{\includegraphics[width=4em]{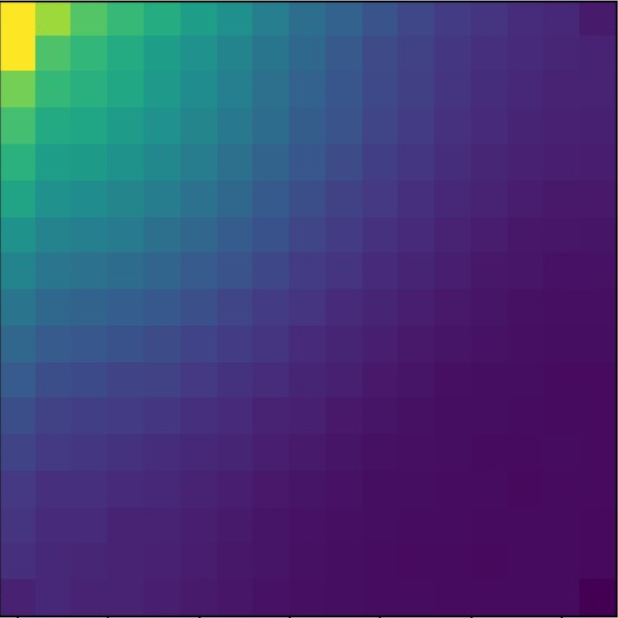}} &
         \parbox[c]{4em}{\includegraphics[width=4em]{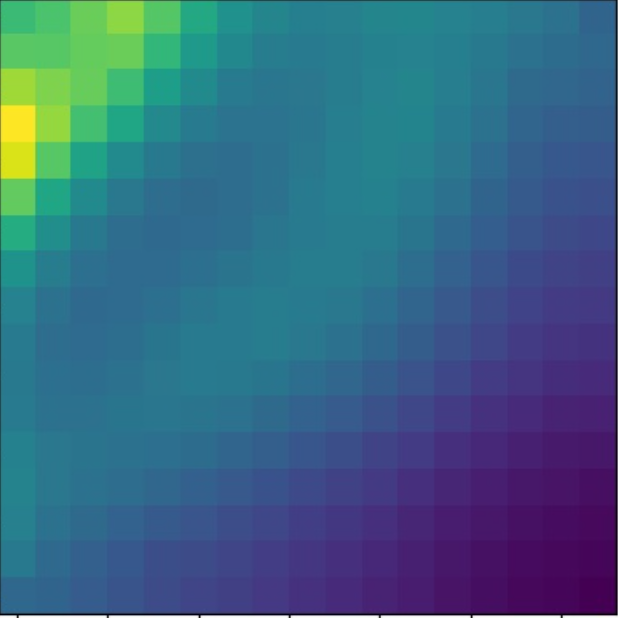}} &
         \parbox[c]{4em}{\includegraphics[width=4em]{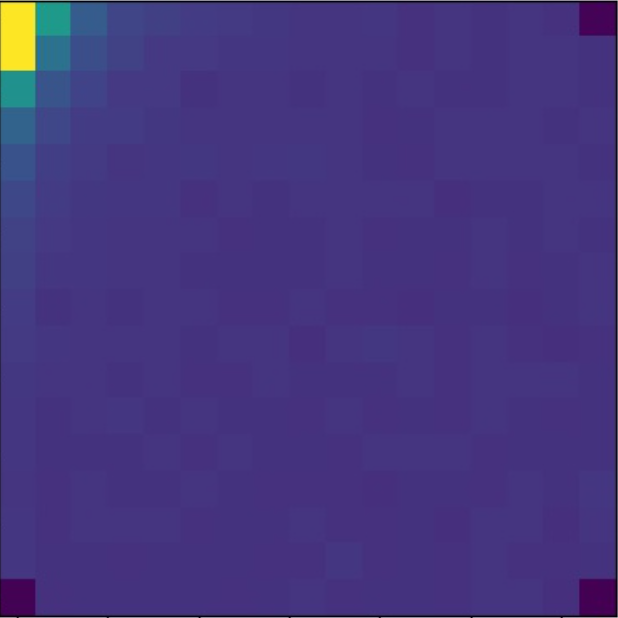}} &
         \parbox[c]{4em}{\includegraphics[width=4em]{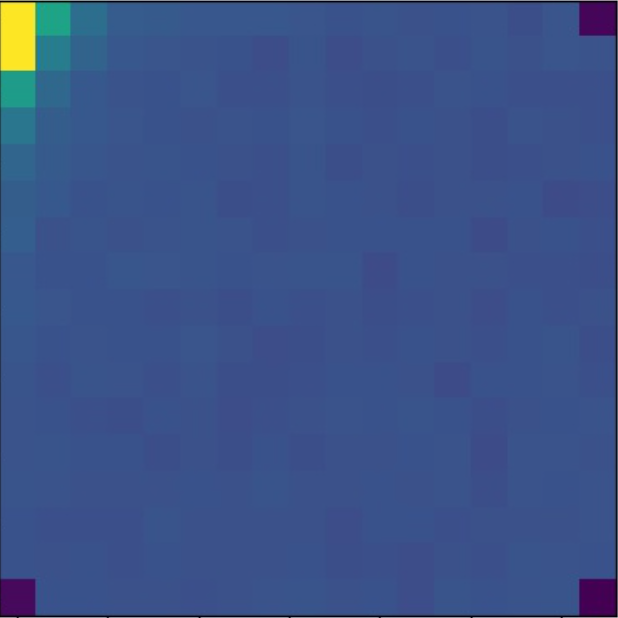}} &
         \parbox[c]{4em}{\includegraphics[width=4em]{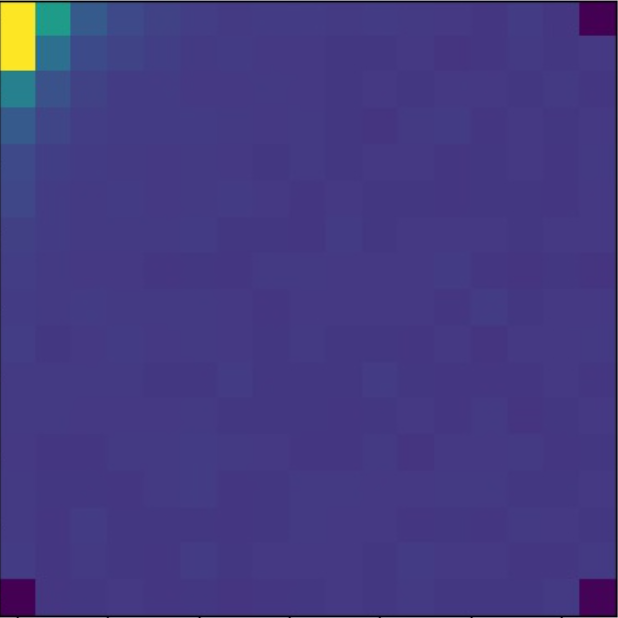}} &
         \parbox[c]{4em}{\includegraphics[width=4em]{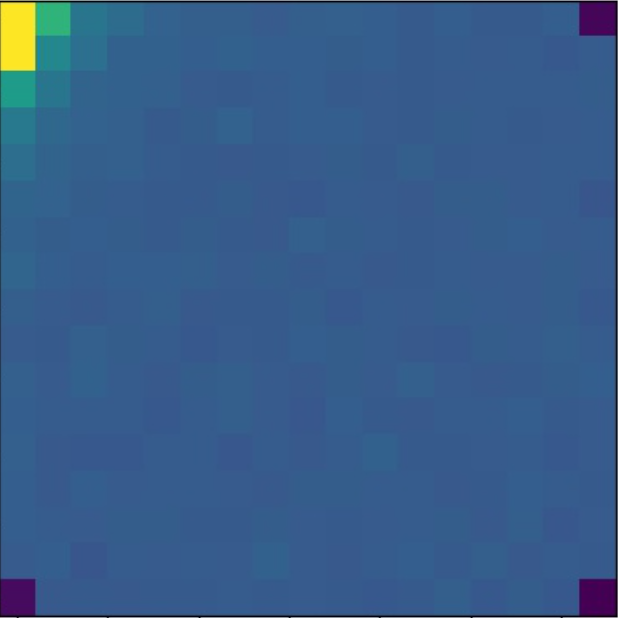}} \\
         \vspace{0.01cm} \\
         $ C(\xinput) $ &          
         \parbox[c]{4em}{\includegraphics[width=4em]{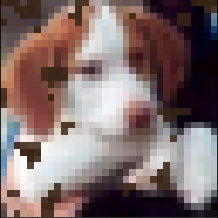}} &
         \parbox[c]{4em}{\includegraphics[width=4em]{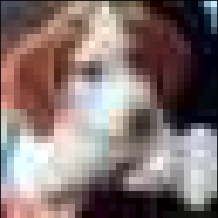}} &
         \parbox[c]{4em}{\includegraphics[width=4em]{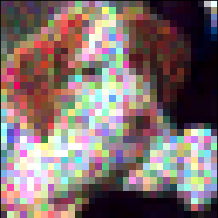}} &
         \parbox[c]{4em}{\includegraphics[width=4em]{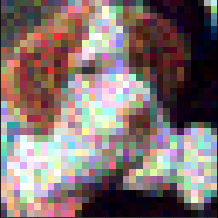}} &
         \parbox[c]{4em}{\includegraphics[width=4em]{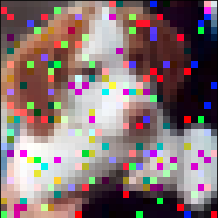}} &
         \parbox[c]{4em}{\includegraphics[width=4em]{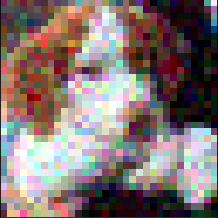}} \\
        \end{tabular}
        
    \caption{CIFAR-10 corruptions arranged according to their frequency bias, i.e., `fog' with largest ratio of low-frequency components indicated by its bigger $\mathcal{B}_{\text{low}}$ and frequency spectrum that shows concentration of intensity at the top-left.}
        
\label{tab:cifar10 corruption stats}
\end{table*}

\end{document}